\documentclass{article}

   \usepackage[nonatbib,final]{neurips_2023}

\usepackage[utf8]{inputenc} 
\usepackage[T1]{fontenc}    
\usepackage{hyperref}       
\usepackage{url}            
\usepackage{booktabs}       
\usepackage{amsfonts}       
\usepackage{nicefrac}       
\usepackage{microtype}      
\usepackage{xcolor}         
\usepackage{graphicx}
\usepackage{subfigure}
\usepackage{parskip}
\usepackage{enumitem}
\usepackage{multirow}
\usepackage{wrapfig}
\usepackage{amsmath}
\usepackage{colortbl}
\usepackage{bbding}
\usepackage[linesnumbered,ruled,vlined]{algorithm2e}
\usepackage{array}

\usepackage{booktabs}

\title{Sparse4D v2

Recurrent Temporal Fusion with Sparse Model}

\author{%
  Xuewu Lin, Tianwei Lin, Zixiang Pei, Lichao Huang, Zhizhong Su \\
  Horizon Robotics, Beijing, China\\
  \texttt{xuewu.lin@horizon.ai} \\
}

\begin{document}
\maketitle
\begin{figure}[htbp]
    \centering
    \subfigure[Multi-frame Sampling and Fusion]{
        \includegraphics[width=0.48\textwidth]{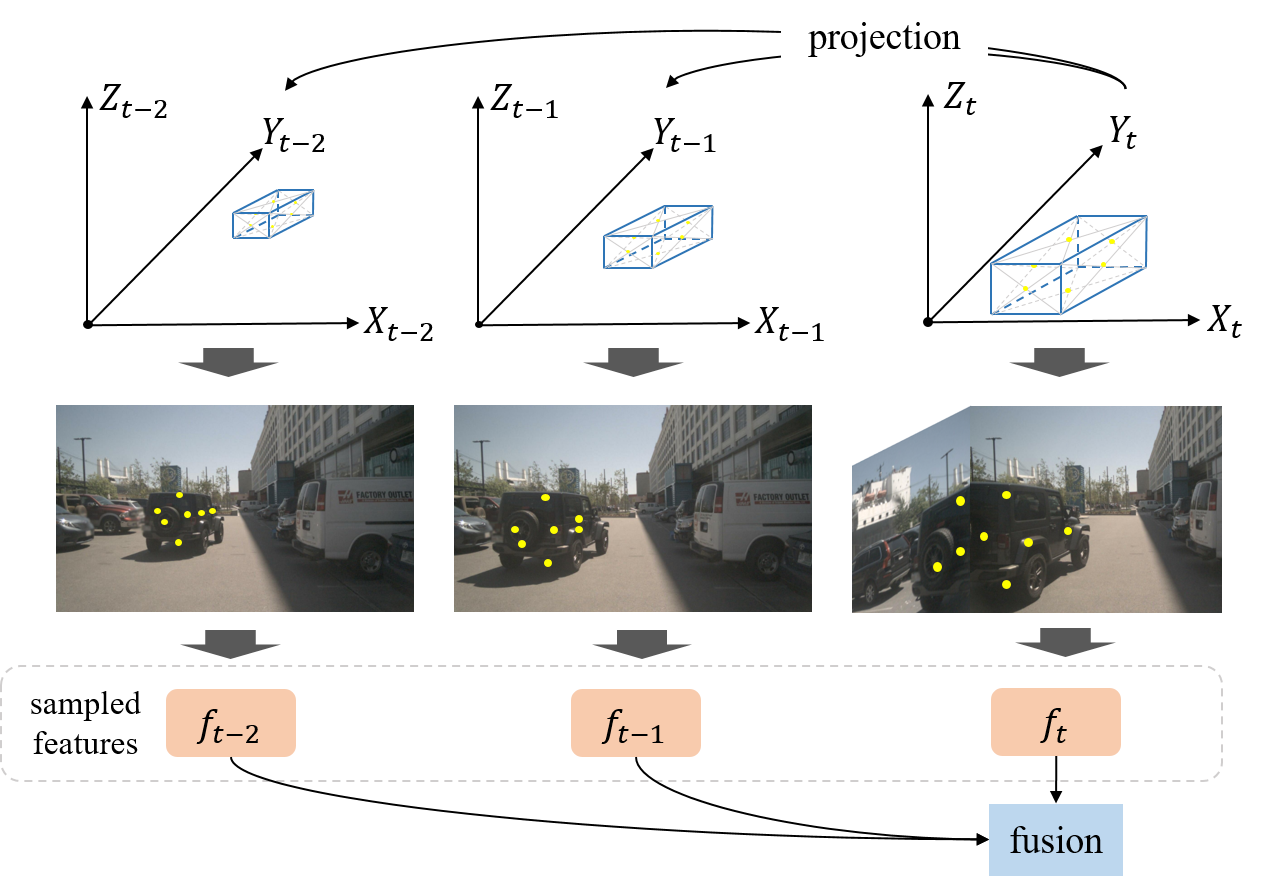}
        \label{multi-frame_sampl}
    }
    \subfigure[Recurrent Temporal Fusion]{
	\includegraphics[width=0.48\textwidth]{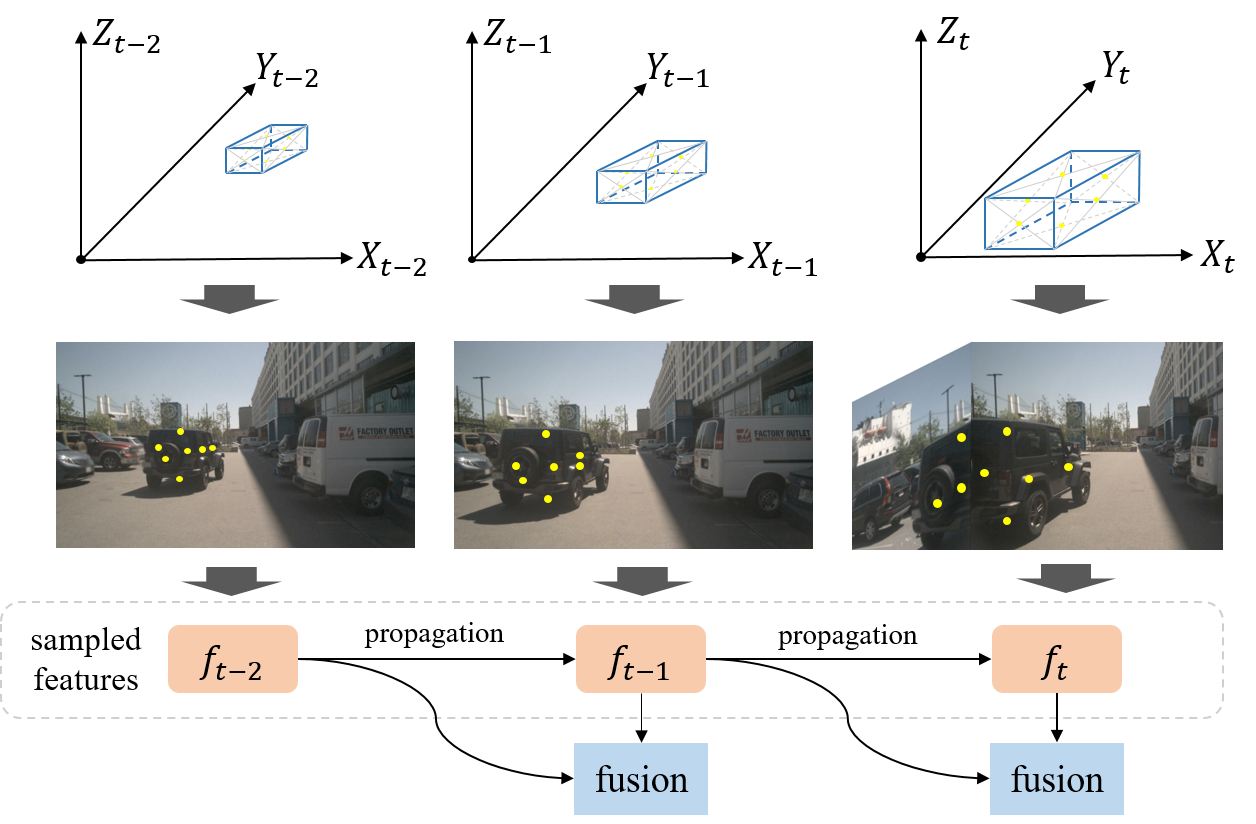}
        \label{Recurrent Temporal Fusion}
    }
    \caption{Comparison of two different temporal fusion approaches. (a) Sparse4D requires projecting the anchors of the current frame onto each historical frame, followed by multi-frame feature sampling and fusion. (b) Sparse4Dv2 achieves fusion through the propagation of instance features.}
    \label{imporvment of v2}
\end{figure}

\begin{abstract}
  Sparse algorithms offer great flexibility for multi-view temporal perception tasks. In this paper, we present an enhanced version of Sparse4D, in which we improve the temporal fusion module by implementing a recursive form of multi-frame feature sampling.
  By effectively decoupling image features and structured anchor features, Sparse4D enables a highly efficient transformation of temporal features, thereby facilitating temporal fusion solely through the frame-by-frame transmission of sparse features.
  The recurrent temporal fusion approach provides two main benefits. Firstly, it reduces the computational complexity of temporal fusion from $O(T)$ to $O(1)$, resulting in significant improvements in inference speed and memory usage. Secondly, it enables the fusion of long-term information, leading to more pronounced performance improvements due to temporal fusion.
  Our proposed approach, Sparse4Dv2, further enhances the performance of the sparse perception algorithm and achieves state-of-the-art results on the nuScenes 3D detection benchmark.
  Code will be available at \url{https://github.com/linxuewu/Sparse4D}.
\end{abstract}

\section{Introduction}
  The perception module plays a critical role in autonomous driving systems. Achieving efficient and effective fusion of multi-sensor and temporal features is a key research direction to enhance perception performance. While bird's-eye-view (BEV) based algorithms~\cite{chen2022bevdistill,li2022bevstereo,solofusion,videobev,li2022bevdepth,huang2021bevdet} have garnered significant attention and demonstrated high perceptual performance, we argue that they may not represent the optimal solution for the following reasons:
  \begin{itemize}[leftmargin=*]
  \item The transformation from image features to the BEV vector space involves reorganizing and rearranging dense features without introducing additional insights. However, this transformation does increase the complexity of the model.
  \item Striking a balance between perception range, accuracy, and computational complexity is crucial for achieving optimal results. Different driving scenarios (e.g., highways, urban or rural areas) require specific parameter settings to ensure an effective trade-off between perceptual capabilities and computational efficiency.
  \item In the context of end-to-end autonomous driving, instance features produced by sparse-based algorithms hold greater significance as they can be more easily integrated with graph-based models like transformers.
  \end{itemize}
  
  Unlike BEV-based algorithms, the PETR series~\cite{liu2022petr,liu2022petrv2,streampetr} algorithms utilize a query-based architecture and global cross-attention to achieve multi-view feature fusion. PETR excludes the dense view-transformation module, but similar to the vanilla DETR~\cite{DETR}, it uses global attention, which results in a high theoretical computational cost. Therefore, it cannot be considered as a purely sparse algorithm. Based on the aforementioned reasons, we remain dedicated to the development of sparse-based algorithms to improve perception performance and prepare for end-to-end autonomous driving. We chose Sparse4D as our baseline algorithm for further enhancements.

  The temporal module of Sparse4D~\cite{lin2022sparse4d} exhibits a notable limitation where it requires sampling multiple frames of historical data before performing feature fusion, as depicted in Figure~\ref{multi-frame_sampl}. This leads to a linear increase in computational complexity with the number of historical frames, resulting in reduced inference and training speed, increased GPU memory usage (as shown in Table~\ref{fps and mem}), and challenges in effectively incorporating long-term temporal features. To overcome this drawback, we propose an alternative solution by replacing the multi-frame sampling approach with a recurrent manner that leverages instance features, similar  to query-based trackers and SteamPETR~\cite{streampetr}, as illustrated in  Figure~\ref{Recurrent Temporal Fusion}.
  
  \begin{table}[h]
  \small
  \caption{Comparison of inference speed and GPU memory usage between sparse4D v1 and v2. Test results based on NVIDIA RTX 3090. Input image size is 704x256, and the backbone is ResNet50.}
  \label{fps and mem}
  \centering
    \begin{tabular}{l|cccccc|c}
      \toprule
      & \multicolumn{6}{c|}{Sparse4Dv1} & Sparse4Dv2  \\
      \midrule
      Frames $T$ & 1  & 2 & 3 & 5 & 7 & 9  & - \\
      FPS & 21.5 & 15.3 & 12.6 & 9.0 & 7.1 & 6.1 & 19.4 \\
      GPU Mem (M) & 424 & 515 & 614 & 792 & 971 & 1149 & 432 \\
      \bottomrule
    \end{tabular}
  \end{table}

  Specifically, for the first frame, we perform detection using single-frame Sparse4D, which outputs a set of 3D bounding boxes along with their corresponding instance features. For subsequent frames, we transform the output from the previous frame to the current frame. The instance features remain unchanged, while the instance states, such as 3D bounding box, are projected onto the current frame as anchors, leveraging the ego motion data. The position embedding of the anchors is also explicitly re-encoded through an anchor encoder. To fully leverage the temporal instance features, we introduce a temporal cross-attention module in each layer of the decoder.
  The instance initialized through temporal projection primarily handle the tracklets, which are objects that have been previously detected and tracked over multiple frames. For newly emerging objects, we initialize them with a single-frame single-layer decoder, selecting the subset of instances with the highest scores to be propagated to the subsequent decoders. This design allows our temporal model to avoid increasing the number of anchors, resulting in comparable inference speed to the non-temporal model.

  In addition to the temporal module, Sparse4Dv2 introduces following improvements: (1) A reconstruction of the deformable aggregation operation by combining bilinear grid sampling and weighted sum into a single CUDA operator, significantly reducing the memory footprint during training and to some extent improving training speed. (2) 
  Incorporating camera parameter encoding into deformable aggregation, along with image and output coordinate augmentation during training, to achieve higher perceptual metrics and robustness. (3) Introducing dense depth supervision based on LiDAR point clouds to facilitate training optimization and enhance the detection accuracy.

  We conducted extensive experiments on the nuScenes 3D detection dataset~\cite{caesar2020nuscenes}, and the results indicate that Sparse4Dv2 exhibits a high level of competitiveness. It outperforms existing BEV-based algorithms such as SOLOFusion~\cite{solofusion} and VideoBEV~\cite{videobev} in terms of perception performance, and also demonstrates an advantage in terms of inference speed.

\section{Related Works}
\subsection{Camera-only 3D Detection}
  The key aspect of camera-only 3D detection tasks is to estimate the depth or 3D coordinates of objects from 2D images. This field primarily encompasses three research directions: monocular 3D, stereo 3D, and multi-view 3D. Monocular 3D is a ill-posed problem that relies on the powerful fitting capability of neural networks to regress the depth in the camera coordinate system by extracting various information from the image~\cite{liu2020smoke,wang2021fcos3d,DD3D,weng2019monocularpseudo}. Stereo 3D involves input from two or more cameras with a significant overlap angle. It utilizes feature point matching and equation-based joint optimization to calculate the depth of target points. Alternatively, depth can be predicted by constructing a cost volume~\cite{mvsnet,fastmvs}. Multi-view 3D lies between monocular 3D and stereo 3D in terms of the percentage of overlap between multiple views. It is a research direction of great interest in the fields of autonomous driving and robotics~\cite{bevformer,huang2022bevdet4d}. It primarily focuses on leveraging the fusion of multi-view and temporal features to enhance depth estimation accuracy.

\subsection{Multi-view Feature Fusion}
  Multi-view feature fusion can significantly improve the perception of objects across different viewpoints. In theory, it has the potential to improve the accuracy of depth estimation in regions where multiple viewpoints overlap. The LSS algorithm~\cite{philion2020lift} utilizes depth estimation results to project image features into the 3D space and performs dense multi-view feature fusion on the BEV plane. When the accuracy of depth estimation is high, LSS-based algorithms~\cite{li2022bevdepth,huang2021bevdet} can achieve improved perception accuracy. The computational complexity of LSS is dependent on the resolution of the input feature maps and the size of the output BEV features. The calculations of LSS is dependent on the resolution of the input feature maps and the size of the output BEV features. Indeed, BEVFormer~\cite{bevformer} also performs feature fusion in the BEV feature space. However, it differs in its approach by employing 3D-to-2D back-projection and utilizing deformable attention for feature fusion. Since BEVFormer needs to output dense BEV features, the query count in the attention can be large (e.g., $200\times200$), which limits the training and inference efficiency. The PETR~\cite{liu2022petr} series abandons the concept of BEV and instead utilizes sparse queries to perform perception tasks. The correspondence between 2D and 3D information is established through 3D position encoding. PETR employs global attention, and the computational efficiency is significantly affected by the resolution of the image features, making it challenging to handle very high-resolution images (e.g., resolutions above 4K). Sparse-based algorithms like DETR3D~\cite{wang2022detr3d}, utilize sparse queries and sparse feature sampling. The computational complexity of the head part is independent of the image resolution, resulting in theoretically high computational efficiency. However, there is a certain gap in performance compared to dense algorithms.

\subsection{Temporal Feature Fusion}
  Temporal feature fusion can greatly improve the location and velocity estimation performance of single-frame models, leading to increased stability in perception results. Initially, temporal fusion was performed using only two frames. BEVDepth~\cite{li2022bevdepth} and BEVDet4D~\cite{huang2022bevdet4d} cache the BEV feature of the previous frame, warp it to the current time step, and concatenate with the feature of the current frame. In PETRv2~\cite{liu2022petrv2}, both the features of the previous and current frames are used as keys for cross attention. Later, it was discovered that long-term temporal feature fusion could further enhance perception performance. Sparse4D~\cite{lin2022sparse4d} utilized 4D key points for feature sampling across multiple frames and then fused them using a fully connected network. SOLOFusion~\cite{solofusion} caches multiple frame features in a silde window, with a window length of up to 16 frames, achieving a substantial improvement in perception performance. VideoBEV~\cite{videobev} transformed the parallel fusion in SOLOFusion into a recurrent form, reducing the computational complexity of fusion. StreamPETR~\cite{streampetr} also adopted a recurrent form, using sparse queries to propagate features across the temporal dimension and achieved state-of-the-art performance.

\section{Methodology}

\begin{figure}
    \centering
    \includegraphics[width=0.9\textwidth]{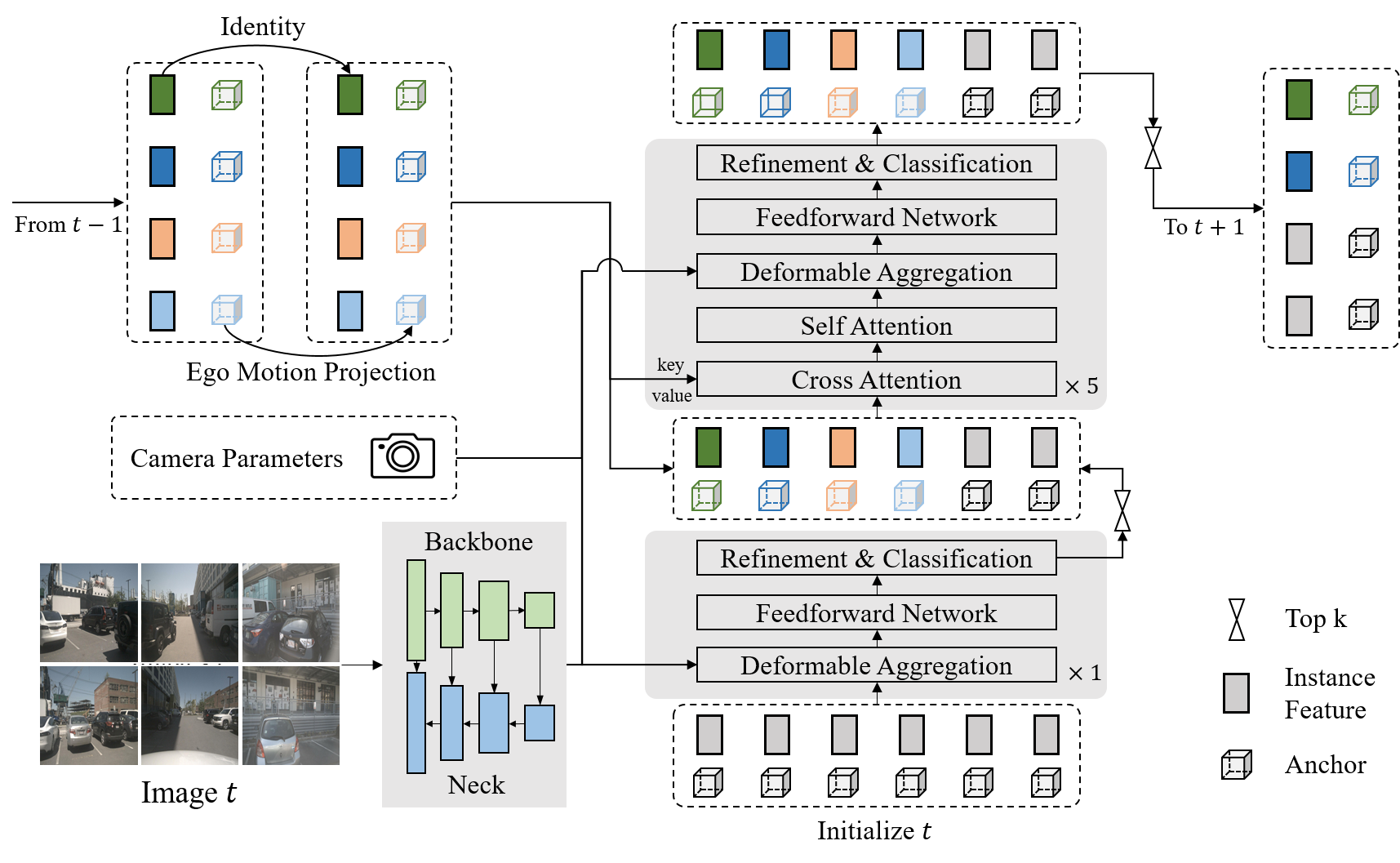}
    \caption{Overall Framework of Sparse4Dv2, which conforms to an encoder-decoder structure. The inputs consists of three components: multi-view images, camera parameters, and instance information from previous frames. The output is the refined instances (anchors and corresponding features), serve as the perception results for the current frame. Additionally, a subset of these instances is selected and used as input for the next frame.
}
    \label{Overall Framework}
\end{figure}

\label{Methodology}
\subsection{Overall Framework}
  As shown in Figure~\ref{Overall Framework}, in Sparse4Dv2, multi-view images are first encoded to extract multi-view multi-scale feature maps $I =\left\{I_{s} \in \mathbb{R}^{N \times C \times H_{s} \times W_{s}} | 1 \le s \le S\right\}$, where $S$ is the number of scales and $N$ is the number of views. These feature maps are then fed into the decoder, which consists of one single-frame layer and five multi-frame layers. The single-frame layer includes three sub-modules: deformable aggregation, feedforward network (FFN), and the output layer for refinement and classification. The multi-frame layers, in addition to the aforementioned sub-modules, also incorporate two multi-head attention layers~\cite{attention}. The cross attention is used to enhance the fusion of temporal features, while the self-attention facilitates feature interaction between instances. First, we initialize a set of instances, which includes their anchor boxes and feature vectors. We refine and score them using a single-frame layer, selecting the highest foreground confidence instances as input to the multi-frame layer. The input instances to the multi-frame layer come not only from the single-frame layer, but also from the outputs of previous frames, such as historical frames. The number of anchors in each layer is consistent, whether it is in the multi-frame or single-frame layer. The output of the multi-frame layer serves as the detection result for the current frame, and a portion of the instances with high confidence scores are selected as input to the next frame.

\subsection{Instance Temporal Propagation}
  In Sparse4D, an instance is represented by three parts, which are anchor, instance feature and anchor embedding. Anchor is structured information, which indicates the state of the instance, and has actual physical meaning. The instance feature is a high-order semantic feature extracted from the image, mainly from the image encoder. And the anchor embedding is the feature encoding of the anchor, and a small anchor encoder $\Psi$ is used to map the structured information of anchor to the high-dimensional space. This design completely decouples the image features and structured state of the instance, so we can add prior knowledge more conveniently. For instance temporal propagation, we only need to project its anchor, and use the anchor encoder to encode the projected anchor, and the instance feature can remain unchanged, the formula is as Equation~\ref{eq:instancetrans}.
    \begin{equation}
    \label{eq:instancetrans}
    A_{t}=\textbf{Project}_{t-1\rightarrow t} (A_{t-1}), \quad E_{t}= \Psi\left(A_{t}\right), \quad F_{t}= F_{t-1}
    \end{equation}
    Where $A$ denotes anchor, $E$ denotes anchor embedding, $F$ is the instance feature and $\Psi$ indicates the anchor encoder. Sparse4D can be used to handle various perception tasks, just need to design different anchor and projection functions for different perception tasks.In particular, for 3D detection, the anchor is defined as a 3D bounding box, and the projection function is as Equation~\ref{eq:box3dproject}.
    \begin{gather}
    A_{t-1}=\left\{x, y, z, w, l, h, \sin{yaw}, \cos{yaw}, v_x, v_y, v_z\right\}_{t-1} \notag\\
    \left[x, y, z\right]_{t} = \textbf{R}_{t-1\rightarrow t} (\left[x, y, z\right] + d_t \left[v_x, v_y, v_z\right])_{t-1} + \textbf{T}_{t-1\rightarrow t} \notag\\
    [w, l, h]_{t} = [w, l, h]_{t-1} \label{eq:box3dproject}\\
    \left[\cos{yaw}, \sin{yaw}, 0\right]_{t} = \textbf{R}_{t-1\rightarrow t} \left[\cos{yaw}, \sin{yaw}, 0\right]_{t-1} \notag\\
    \left[v_x, v_y, v_z\right]_{t} = \textbf{R}_{t-1\rightarrow t} \left[v_x, v_y, v_z\right]_{t-1} \notag
    \end{gather}
    
    where $d_t$ is the time interval between frame $t$ and $t-1$, $\textbf{R}_{t-1\rightarrow t}$ and $\textbf{T}_{t-1\rightarrow t}$ represent the rotation matrix and translation of the ego vehicle from time step $t-1$ to $t$, respectively. If 3D lane line detection is desired, the anchors can be defined as polylines, and the projection function would be the projection of each 3D point on the polyline.

\subsection{Efficient Deformable Aggregation}
  The Deformable Aggregation is designed to achieve feature fusion across multiple scales and views. It involves feature sampling from multiple feature maps and weighted summation. One of the most fundamental implementations is shown in Algorithm~\ref{basicdeformable}.
\IncMargin{1em}
\begin{algorithm} \SetKwData{Left}{left}\SetKwData{This}{this}\SetKwData{Up}{up} \SetKwFunction{Union}{Union}\SetKwFunction{FindCompress}{FindCompress} \SetKwInOut{Input}{input}\SetKwInOut{Output}{output}
    \Input{
        \textbf{1) feature maps} $I =\left\{I_{s} \in \mathbb{R}^{N \times C \times H_{s} \times W_{s}} | 1 \le s \le S\right\}$, \textbf{2) projected 2D points} $P \in \mathbb{R}^{K \times N \times 2}$, \textbf{3) weights} $W \in \mathbb{R}^{K \times N \times S \times G}$. $C$ is the feature channels and $K$ is the number of points. 
    }
    \Output{
        features $F \in \mathbb{R}^{C}$}
    \BlankLine 

    All inputs are stored in high bandwidth memory (HBM)\;
    Initializes an empty list $f$\;
    \For{$i\leftarrow 1$ \KwTo $S$}{
      Write \textbf{Blinear}$(I_{s}, P) \in \mathbb{R}^{N \times C \times K}$ to HBM and add to $f$;
    }
    Stack and reshape $f$ to the shape can be broadcast to $W$, and write to HBM\;
    Multiply sampled features and weights $f=f\times W \in \mathbb{R}^{K\times N \times S \times C}$, and write to HBM\;
    Sum $f$ along the view/scale/point dimensions and write it to HBM as output $F$.
\caption{Basic Deformable Aggregation}
\label{basicdeformable} 
\end{algorithm}
\DecMargin{1em}

It can be observed that Algorithm~\ref{basicdeformable} requires multiple loads and stores of HBM. It also
\begin{wrapfigure}{l}{0.5\textwidth}
    \centering
    \includegraphics[width=0.5\textwidth]{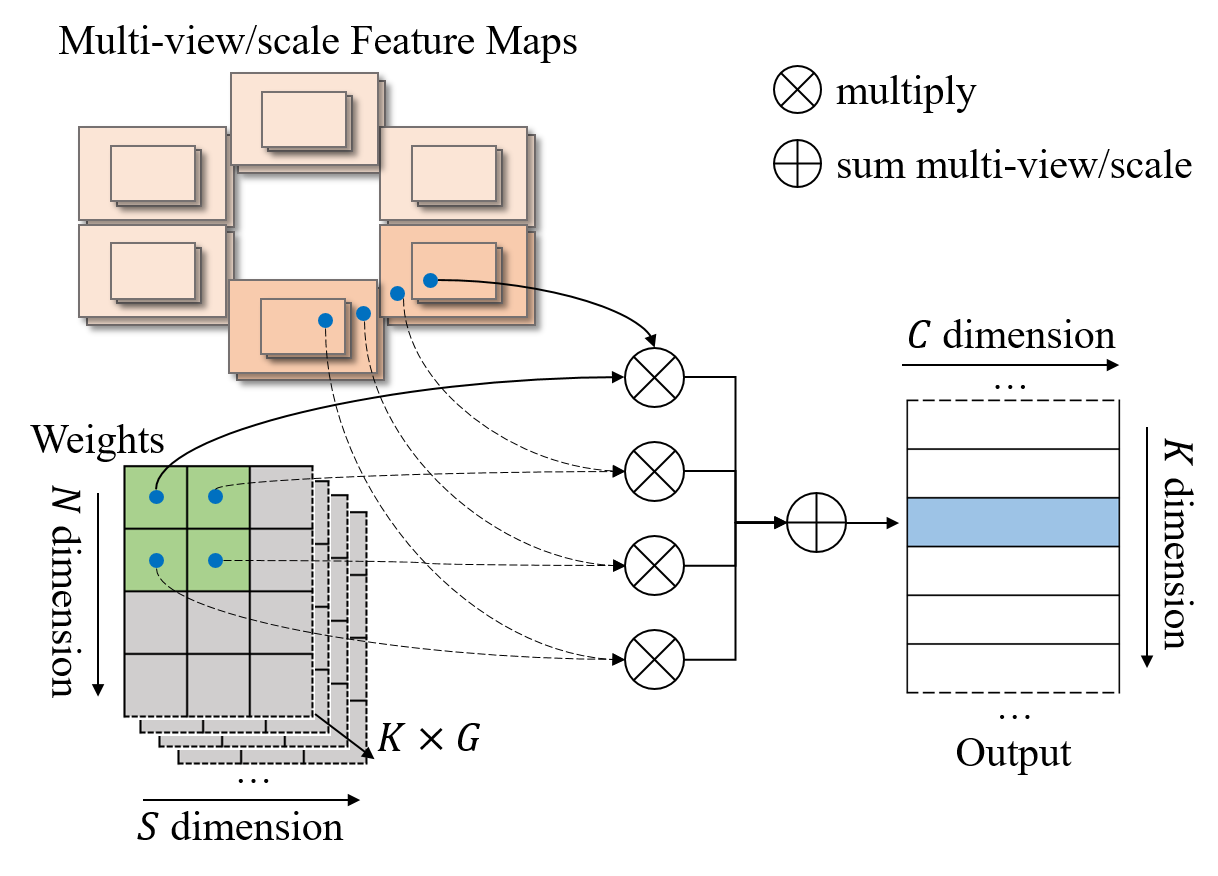}
    \caption{Efficient Deformable Aggregation.}
    \label{deformableagg}
\end{wrapfigure}
   requires storing numerous intermediate variables for backpropagation during training, especially when the size of $f$ is large. As a result, it consumes a significant amount of GPU memory. Additionally, the frequent HBM access also slows down the inference speed. To address this problem, we encapsulate the feature sampling and scale/view dimension weighting in the Basic Deformable Aggregation as a CUDA operation, directly outputting multi-point features in a single step. We refer to this optimization as Efficient Deformable Aggregation (EDA), as shown in Figure~\ref{deformableagg}.
EDA exhibits excellent parallel performance, allowing for complete parallelization in the $K$ and $C$ dimensions. The computational complexity for a single thread is only $N\times S$. Additionally, for multi-view scenarios, a point is projected to at most two views, resulting in a single-thread computational complexity of only $2S$. This design leverages the parallel computing capabilities of GPUs and AI chips, significantly improving efficiency and reducing both memory consumption and inference time. EDA can serve as a versatile op, suitable for various applications requiring multi-image and multi-scale fusion.

\subsection{Camera Parameter Encoding}
  In Sparse4Dv1, the weights in Deformable Aggregation are computed through a fully connected layer. The information of camera parameters is gradually embedded into the parameters of this fully connected layer during the training process, which can be seen as an implicit neural representation or as overfitting to the training set. This approach can lead to poor generalization to camera parameters. Specifically, when swapping the input order of two images, the order of weights does not change accordingly, which can impact perception performance. Additionally, if we apply large-scale data augmentation on camera parameters during training, the convergence speed of this parameterized implicit representation may be significantly affected. To address this, we directly input camera parameters into the network and map the transformation matrix from output space to image coordinate space into a high-dimensional feature vector. We then add this feature vector to the instance feature and use the combined feature to calculate the weights for the corresponding view.

\subsection{Dense Depth Supervision}
  During our experiments, we observed that sparse-based methods lacked sufficient convergence capability and speed in the early training stage. To alleviate this problem, we introduced multi-scale dense depth estimation with point clouds as supervision. During inference, this sub-network will not be activated. For each scale of the feature map, we employ a $1\times 1$ convolution to output depth values at a pre-defined equivalent focal length. These depth values are then multiplied by the ratio of the camera focal length to the  equivalent focal length. The loss function for dense depth supervision uses the vanilla L1 loss. After incorporating dense depth supervision, we will remove the depth-reweight module from Sparse4Dv1.

\section{Experiment}
\label{Experiment}
\subsection{Benckmark}
  In order to validate the effectiveness of Sparse4Dv2, we utilized the nuScenes benchmark. The dataset consists of 1000 scenes, with 700 scenes for training, 150 for validation, and 150 for testing. Each scene comprises a 20-second video clip captured at 2 frames per second (fps), with 6 viewpoint images.  In addition to the 3D bounding box labels for 10 different object classes, the dataset also provides information on the vehicle motion states and camera parameters. The evaluation metrics include mean Average Precision (mAP), mean Average Error of Translation (mATE), Scale (mASE), Orientation (mAOE), Velocity (mAVE), Attribute (mAAE) and nuScenes Detection Score (NDS), where NDS is a weighted average of other metrics. Please refer \cite{caesar2020nuscenes} for details.
  
  Due to our goal of improving the performance of sparse-based algorithms, our direct baseline for comparison is Sparse4Dv1. Additionally, to showcase the competitiveness of sparse-based algorithms, we will also compare our method against other state-of-the-art BEV-based algorithms.

\subsection{Implementation Details}
  We utilized ResNet50, ResNet101~\cite{resnet}, and VoVNet-99~\cite{vovnet} as backbones, along with FPN~\cite{fpn} as the neck architecture. The pretrained weights for ResNet models were obtained from ImageNet-1K~\cite{krizhevsky2017imagenet} and nuImages datasets, while the pretrained weights for VoVNet were provided by DD3D~\cite{DD3D}. All experiments were trained for 100 epochs using the AdamW optimizer~\cite{adamw} without CBGS~\cite{zhu2019cbgs}. Similar to most open-source algorithms, we employed image data augmentation and lidar rotation augmentation techniques. To improve training efficiency for temporal models, we adopted the sequential iteration approach inspired by SOLOFusion~\cite{solofusion}. We utilized a total of 900 instance anchors, with 600 of them being temporal instances from the historical output, and the remaining 300 coming from the single-frame layer. The remaining hyperparameters for network are kept consistent with Sparse4Dv1~\cite{lin2022sparse4d}.

\subsection{Ablation Studies and Analysis}
  In our ablation experiments, we consistently used ResNet50 with ImageNet-1K pretraining. The input image size was set to $256\times 704$.
  
  \textbf{Efficient Deformable Aggregation} significantly impacts GPU memory usage and inference speed, as shown in Table~\ref{tab:eda}.During the training phase, when the batch size is set to 1, the GPU memory consumption decreases from 6328 MB to 3100 MB, resulting in a 51\% reduction. The maximum available batch size increases from 3 to 8. Additionally, the total training time for a complete experiment is reduced from 23.5 hours to 14.5 hours. These improvements greatly reduce the training barrier and enhance the efficiency of Sparse4D. In the inference phase, EDA improves the model's FPS from 13.7 to 20.3, resulting in a significant improvement of approximately 42\%. Furthermore, EDA reduces GPU memory usage during inference by 53\%, which lowers the cost and deployment complexity of Sparse4D.
  \begin{table*}[h]
  \centering
  \small
  \setlength{\tabcolsep}{1.5mm}{
  \begin{tabular}{@{}c|c|c|c|c|c@{}}
    \toprule
    & \multicolumn{3}{c|}{Training} & \multicolumn{2}{c}{Inference}\\
    EDA & GPU Memory (M)& Max Batch Size & Time (h) & GPU Memory (M) & FPS \\
    \midrule
     & 6328& 3 & 23.5& 925 & 13.7 \\
    \checkmark & 3100 & 8 & 14.5& 432 & 20.3 \\
    \bottomrule
  \end{tabular}}
  \caption{Ablation of Efficient Deformable Aggregation. This experiment was conducted using RTX 3090 GPU with 24 GB memory. For measurements of Training GPU Memory, Infer GPU Memory, and Infer FPS, we used a batch size of 1. The Training Time represents the time taken to train the model for 100 epochs using the maximum batch size on 8 RTX 3090 GPUs.}
  \label{tab:eda}
\end{table*}

  \textbf{Camera Parameter Encoding} explicitly incorporates camera parameters into the network. From the metrics, it can be observed that removing this module results in a decrease of 2.0 mAP and 4.8 mAOE, as shown in Table~\ref{tab:ablation} (Exp.3 and Exp.5),. In theory, estimating the orientation of an object not only relies on its image features but also requires knowledge of the object's position in the coordinate system and camera intrinsic and extrinsic parameters. Therefore, the explicit encoding of camera parameters can enhance the accuracy of orientation estimation.

  \textbf{Dense Depth Supervision} is mainly to make Sparse4D easier to train. Compared to Sparse4Dv1, we added more data augmentation and changed the pretraining parameters from FCOS3d to ImageNet, making the model training more challenging. Comparing Exp.4 and Exp.5 in Table~\ref{tab:ablation}, it can be observed that removing Dense Depth Supervision leads to a significant decrease in performance metrics (mAP and NDS decrease by 8.5 and 10.4, respectively).  This decrease is attributed to the occurrence of gradient collapse during the training process.
  
  The temporal decoder consists of a \textbf{Single-Frame Layer} and five \textbf{Multi-Frame Layers}. The single-frame layer is primarily designed to handle the selection of newly generated objects. If all layers in the decoder are changed to multi-frame layers, the number of anchors dedicated to handling new objects will decrease while maintaining the predetermined total anchor quantity. This reduction in the number of anchors for new objects leads to a decrease in detection results, as shown in  Exp.2 and Exp.3 in Table~\ref{tab:ablation} (with a decrease of 3.5 in mAP and 2.0 in NDS, respectively). If the multi-frame layers are removed, and all layers in the decoder receive input instances solely from the current frame, it will result in a non-temporal fusion model. 
  Comparing the non-temporal model to the temporal model (Exp.1 and Exp.5 in Table~\ref{tab:ablation}) directly showcases the significant impact of temporal fusion. It brings about a significant improvement of 9.8 mAP and 12.5 NDS.

\begin{table*}[h]
  \centering
  \small
  \setlength{\tabcolsep}{1.5mm}{
  \begin{tabular}{@{}l|cccc|cccccc|c@{}}
    \toprule
    Exp & MF & SFL  & CPE & DDS & mAP$\uparrow$ & mATE$\downarrow$ & mASE$\downarrow$ & mAOE$\downarrow$ & mAVE$\downarrow$ & mAAE$\downarrow$ & NDS$\uparrow$ \\
    \midrule
     1 &  &  - & \checkmark &\checkmark & 0.341 & 0.710 & 0.277 & 0.544 & 0.803 & 0.227 & 0.414\\
    2 & \checkmark &  &  & \checkmark & 0.384 & 0.621 & 0.271 & 0.524 & \textbf{0.281} & 0.180 & 0.504 \\
    3 & \checkmark & \checkmark &  & \checkmark &  0.419 & 0.594 & \textbf{0.268} & 0.523 & 0.285 & 0.203 & 0.524 \\
    4* &\checkmark & \checkmark & \checkmark &  & 0.354 &  0.680 & 0.283 & 0.730 & 0.515 & 0.208 & 0.435 \\
    5&\checkmark & \checkmark & \checkmark & \checkmark & \textbf{0.439} & \textbf{0.598} & 0.270 & \textbf{0.475} & 0.282 & \textbf{0.179} & \textbf{0.539} \\
    \bottomrule
  \end{tabular}}
  \caption{Ablation Experiments. MF (Multi-Frame), SFL (Single-Frame Layer), CPE (Camera Parameter Encoding), and DDS (Dense Depth Supervision). * of Exp.4 indicates that the experiment failed during the training process due to training instability.}
  \label{tab:ablation}
\end{table*}

\subsection{Main Results}
  To better control variables, we first compare our method with other approaches on the nuScenes validation dataset, and the results are shown in Table~\ref{tab:detectionval}. Under the setting of ResNet50 and low-resolution input, Sparse4Dv2 achieves the best mAP and NDS, surpassing the SOTA BEV-based algorithm VideoBEV by 1.7 mAP and 0.4 NDS, and outperforming the query-based SOTA algorithm StreamPETR by 0.7 mAP and 0.2 NDS. Under the setting of ResNet101 and high-resolution input, Sparse4Dv2 also achieves the highest performance metrics. Compared to the baseline Sparse4Dv1, the revised version improves NDS by 3.0 and inference speed by 2.9 times. Compared to StreamPETR, Sparse4Dv2 also demonstrates advantages with a 0.2 NDS improvement and 2 FPS higher inference speed. When we introduce an additional future frame, Sparse4Dv2 still maintains faster inference speed than StreamPETR, and achieves a 1.6 higher NDS score.
  
  It is important to note that when low-resolution images are used as input, Sparse4Dv2 exhibits significantly lower inference speed compared to StreamPETR (20.3 vs 26.7 FPS). However, when the resolution is increased to $512\times 1408$, Sparse4Dv2 surpasses StreamPETR in terms of speed (8.4 vs 6.4 FPS). The computational workload of the Sparse4Dv2 head is independent of the input resolution, making it \textbf{more suitable for high-resolution applications such as long-range detection}. This further demonstrates the advantages of the sparse decoder.

  Table~\ref{tab:detectiontest} presents the performance metrics on the nuScenes test dataset. Comparing to Sparse4Dv1, the enhanced version in this paper has improved all metrics and achieved a significant advancement of 4.3 NDS. In terms of the metric NDS, we have achieved state-of-the-art (SOTA) performance, surpassing SOLOFusion, BEVFormerv2, VideoBEV, and StreamPETR.

\begin{table*}
  \centering
  \scriptsize
  \setlength{\tabcolsep}{1.5mm}{
  \begin{tabular}{@{}l|c|c|cccccc|c|c@{}}
    \toprule
    Method & Backbone & Image size & mAP$\uparrow$ & mATE$\downarrow$ & mASE$\downarrow$ & mAOE$\downarrow$ & mAVE$\downarrow$ & mAAE$\downarrow$ & NDS$\uparrow$ & FPS$\uparrow$ \\
    \midrule
    BEVPoolv2~\cite{huang2022bevpoolv2} & ResNet50 & $256\times 704$ & 0.406 & 0.572 & 0.275 & 0.463 & 0.275 & 0.188 & 0.526 & 16.6 \\
    BEVFormerV2~\cite{bevformerv2} & ResNet50 & - & 0.423 & 0.618 & 0.273 & \textbf{0.413} & 0.333 & 0.188 & 0.529 & - \\
    SOLOFusion~\cite{solofusion} & ResNet50 & $256\times 704$ & 0.427 & 0.567 & 0.274 & 0.511 & \textbf{0.252} & 0.181 & 0.534 & 11.4 \\
    VideoBEV~\cite{videobev} & ResNet50 & $256\times 704$ & 0.422 & \textbf{0.564} & 0.276 & 0.440 & 0.286 & 0.198 & 0.535 & -\\
    StreamPETR~\cite{streampetr} & ResNet50 & $256\times 704$ & 0.432 & 0.609 & \textbf{0.270} & 0.445 & 0.279 & 0.189 & 0.537 & \textbf{26.7} \\
    \specialrule{0em}{1pt}{1pt}
    \rowcolor{gray!30} Sparse4Dv2 & ResNet50 & $256\times 704$ & \textbf{0.439} & 0.598 & \textbf{0.270} & 0.475 & 0.282 & \textbf{0.179} & \textbf{0.539} & 20.3 \\
    \specialrule{0em}{2pt}{2pt}
    \bottomrule
    \specialrule{0em}{2pt}{2pt}
    
    BEVDepth~\cite{li2022bevdepth} & ResNet101 & $512\times 1408$ & 0.412 & 0.565 & 0.266 & 0.358 & 0.331 & 0.190 & 0.535 & - \\
    Sparse4D~\cite{lin2022sparse4d} & Res101-DCN & $640\times 1600$ & 0.444 & 0.603 & 0.276 & 0.360 & 0.309 & \textbf{0.178} & 0.550 & 2.9 \\
    SOLOFusion & ResNet101 & $512\times 1408$ & 0.483 & \textbf{0.503} & 0.264 & 0.381 & 0.246 &  0.207 & 0.582 & - \\
    StreamPETR$^{\dagger}$ & ResNet101 & $512\times 1408$ & 0.504 & 0.569 & \textbf{0.262} & \textbf{0.315} & 0.257 & 0.199 & 0.592 & 6.4 \\
    
    \specialrule{0em}{1pt}{1pt}
    \rowcolor{gray!30} Sparse4Dv2 & ResNet101 & $512\times 1408$ & 0.485 & 0.555 & 0.272 & 0.367 & 0.256 & 0.182 & 0.580 & \textbf{8.4} \\
    \rowcolor{gray!30} Sparse4Dv2$^{\dagger}$ & ResNet101 & $512\times 1408$ & 0.505 & 0.548 & 0.268 & 0.348 & 0.239 & 0.184 & 0.594 & \textbf{8.4} \\
    \rowcolor{gray!30} Sparse4Dv2$^{\dagger *}$ & ResNet101 & $512\times 1408$ & \textbf{0.521} &  0.519 & 0.265 & 0.364 & \textbf{0.199} & 0.180 & \textbf{0.608} & 7.1 \\
    \bottomrule
  \end{tabular}}
  \caption{Results of 3D detection on nuScenes validation dataset. $\dagger$ indicates to use pre-trained weights from the nuImage dataset, and * means to use a future frame.}
  \label{tab:detectionval}
  \vspace{-0.3cm}
\end{table*}

\begin{table*}
  \centering
  \scriptsize
  \setlength{\tabcolsep}{1.5mm}{
  \begin{tabular}{@{}l|c|cccccc|c@{}}
    \toprule
    Method & Backbone & mAP$\uparrow$ & mATE$\downarrow$ & mASE$\downarrow$ & mAOE$\downarrow$ & mAVE$\downarrow$ & mAAE$\downarrow$ & NDS$\uparrow$ \\
    \midrule
    Sparse4D & VovNet-99 & 0.511 & 0.533 & 0.263 & 0.369 & 0.317 & 0.124 & 0.595\\
    HoP-BEVFormer~\cite{hop} & VovNet-99 & 0.517 & 0.501 & 0.245 & 0.346 & 0.362 & \textbf{0.105} & 0.603 \\ 
    SOLOFusion & ConvNeXt-B~\cite{liu2022convnet} & 0.540 & \textbf{0.453} & 0.257 & 0.376 & 0.276 & 0.148 & 0.619 \\
    BEVFormerv2 & InternImage-B~\cite{wang2022internimage} & 0.540 & 0.488 & 0.251 & 0.335 & 0.302 & 0.122 & 0.620 \\
    VideoBEV & ConvNeXt-B & 0.554 & 0.457 & 0.249 & 0.381 & 0.266 & 0.132 & 0.629 \\
    BEVFormerv2 & InternImage-XL & 0.556 & 0.456 & 0.248 & 0.317 & 0.293 & 0.123 & 0.634 \\
    StreamPETR & VovNet-99 & 0.550 & 0.479 & 0.239 & \textbf{0.317} & \textbf{0.241} & 0.119 & 0.636 \\
    \specialrule{0em}{1pt}{1pt}
    \rowcolor{gray!30} Sparse4Dv2 & VovNet-99 & \textbf{0.557} & 0.462 & \textbf{0.238} & 0.328 & 0.264 & 0.115 & \textbf{0.638}
 \\
    \bottomrule
  \end{tabular}}
  \caption{Results of 3D detection on nuScenes test dataset.}
  \label{tab:detectiontest}
  \vspace{-0.3cm}
\end{table*}

\section{Conclusion and Outlook}
\label{Conclusion}
In this paper, we focuses on enhancing the performance of sparse-based algorithms for multi-view temporal perception. Based on Sparse4D, a series of improvements were made, including:
1) Structure: The temporal module was transformed from a multi-frame sampling approach to a recurrent manner, and camera-parameter encoding was incorporated.
2) Training optimization: Dense depth supervision was introduced as an auxiliary supervision signal to improve training performance.
3) Efficiency: Efficient Deformable Aggregation was proposed to reduce training memory usage and improve training and inference speed.
Experiments were conducted on the nuScenes 3D detection dataset. The results demonstrate that our improved model, Sparse4Dv2, not only achieved significant improvements in inference speed but also made substantial advancements in detection performance, reaching SOTA levels.

Further exploration is needed to validate Sparse4Dv2, including its generalization, scene robustness, and long-range detection performance. Additionally, there is still significant research potential for sparse-based methods in various areas such as HD map construction, topology, trajectory prediction, and end-to-end planning. In future studies, we hope that Sparse4Dv2 can serve as a new baseline and be further investigated in these directions.

{
\bibliographystyle{ieee_fullname}
\bibliography{egbib}

\begin{thebibliography}{10}\itemsep=-1pt

\bibitem{caesar2020nuscenes}
Holger Caesar, Varun Bankiti, Alex~H Lang, Sourabh Vora, Venice~Erin Liong,
  Qiang Xu, Anush Krishnan, Yu Pan, Giancarlo Baldan, and Oscar Beijbom.
\newblock nuscenes: A multimodal dataset for autonomous driving.
\newblock In {\em Proceedings of the IEEE/CVF conference on computer vision and
  pattern recognition}, pages 11621--11631, 2020.

\bibitem{DETR}
Nicolas Carion, Francisco Massa, Gabriel Synnaeve, Nicolas Usunier, Alexander
  Kirillov, and Sergey Zagoruyko.
\newblock End-to-end object detection with transformers.
\newblock In {\em European conference on computer vision}, pages 213--229.
  Springer, 2020.

\bibitem{chen2022bevdistill}
Zehui Chen, Zhenyu Li, Shiquan Zhang, Liangji Fang, Qinhong Jiang, and Feng
  Zhao.
\newblock Bevdistill: Cross-modal bev distillation for multi-view 3d object
  detection.
\newblock {\em arXiv preprint arXiv:2211.09386}, 2022.

\bibitem{videobev}
Chunrui Han, Jianjian Sun, Zheng Ge, Jinrong Yang, Runpei Dong, Hongyu Zhou,
  Weixin Mao, Yuang Peng, and Xiangyu Zhang.
\newblock Exploring recurrent long-term temporal fusion for multi-view 3d
  perception.
\newblock {\em arXiv preprint arXiv:2303.05970}, 2023.

\bibitem{resnet}
Kaiming He, Xiangyu Zhang, Shaoqing Ren, and Jian Sun.
\newblock Deep residual learning for image recognition.
\newblock In {\em Proceedings of the IEEE conference on computer vision and
  pattern recognition}, pages 770--778, 2016.

\bibitem{huang2022bevdet4d}
Junjie Huang and Guan Huang.
\newblock Bevdet4d: Exploit temporal cues in multi-camera 3d object detection.
\newblock {\em arXiv preprint arXiv:2203.17054}, 2022.

\bibitem{huang2022bevpoolv2}
Junjie Huang and Guan Huang.
\newblock Bevpoolv2: A cutting-edge implementation of bevdet toward deployment.
\newblock {\em arXiv preprint arXiv:2211.17111}, 2022.

\bibitem{huang2021bevdet}
Junjie Huang, Guan Huang, Zheng Zhu, and Dalong Du.
\newblock Bevdet: High-performance multi-camera 3d object detection in
  bird-eye-view.
\newblock {\em arXiv preprint arXiv:2112.11790}, 2021.

\bibitem{krizhevsky2017imagenet}
Alex Krizhevsky, Ilya Sutskever, and Geoffrey~E Hinton.
\newblock Imagenet classification with deep convolutional neural networks.
\newblock {\em Communications of the ACM}, 60(6):84--90, 2017.

\bibitem{vovnet}
Youngwan Lee, Joong-won Hwang, Sangrok Lee, Yuseok Bae, and Jongyoul Park.
\newblock An energy and gpu-computation efficient backbone network for
  real-time object detection.
\newblock In {\em Proceedings of the IEEE/CVF conference on computer vision and
  pattern recognition workshops}, pages 0--0, 2019.

\bibitem{li2022bevstereo}
Yinhao Li, Han Bao, Zheng Ge, Jinrong Yang, Jianjian Sun, and Zeming Li.
\newblock Bevstereo: Enhancing depth estimation in multi-view 3d object
  detection with dynamic temporal stereo.
\newblock {\em arXiv preprint arXiv:2209.10248}, 2022.

\bibitem{li2022bevdepth}
Yinhao Li, Zheng Ge, Guanyi Yu, Jinrong Yang, Zengran Wang, Yukang Shi,
  Jianjian Sun, and Zeming Li.
\newblock Bevdepth: Acquisition of reliable depth for multi-view 3d object
  detection.
\newblock {\em arXiv preprint arXiv:2206.10092}, 2022.

\bibitem{bevformer}
Zhiqi Li, Wenhai Wang, Hongyang Li, Enze Xie, Chonghao Sima, Tong Lu, Qiao Yu,
  and Jifeng Dai.
\newblock Bevformer: Learning bird's-eye-view representation from multi-camera
  images via spatiotemporal transformers.
\newblock {\em arXiv preprint arXiv:2203.17270}, 2022.

\bibitem{fpn}
Tsung-Yi Lin, Piotr Doll{\'a}r, Ross Girshick, Kaiming He, Bharath Hariharan,
  and Serge Belongie.
\newblock Feature pyramid networks for object detection.
\newblock In {\em Proceedings of the IEEE conference on computer vision and
  pattern recognition}, pages 2117--2125, 2017.

\bibitem{lin2022sparse4d}
Xuewu Lin, Tianwei Lin, Zixiang Pei, Lichao Huang, and Zhizhong Su.
\newblock Sparse4d: Multi-view 3d object detection with sparse spatial-temporal
  fusion.
\newblock {\em arXiv preprint arXiv:2211.10581}, 2022.

\bibitem{liu2022petr}
Yingfei Liu, Tiancai Wang, Xiangyu Zhang, and Jian Sun.
\newblock Petr: Position embedding transformation for multi-view 3d object
  detection.
\newblock {\em arXiv preprint arXiv:2203.05625}, 2022.

\bibitem{liu2022petrv2}
Yingfei Liu, Junjie Yan, Fan Jia, Shuailin Li, Qi Gao, Tiancai Wang, Xiangyu
  Zhang, and Jian Sun.
\newblock Petrv2: A unified framework for 3d perception from multi-camera
  images.
\newblock {\em arXiv preprint arXiv:2206.01256}, 2022.

\bibitem{liu2022convnet}
Zhuang Liu, Hanzi Mao, Chao-Yuan Wu, Christoph Feichtenhofer, Trevor Darrell,
  and Saining Xie.
\newblock A convnet for the 2020s.
\newblock In {\em Proceedings of the IEEE/CVF Conference on Computer Vision and
  Pattern Recognition}, pages 11976--11986, 2022.

\bibitem{liu2020smoke}
Zechen Liu, Zizhang Wu, and Roland T{\'o}th.
\newblock Smoke: Single-stage monocular 3d object detection via keypoint
  estimation.
\newblock In {\em Proceedings of the IEEE/CVF Conference on Computer Vision and
  Pattern Recognition Workshops}, pages 996--997, 2020.

\bibitem{adamw}
Ilya Loshchilov and Frank Hutter.
\newblock Decoupled weight decay regularization.
\newblock {\em arXiv preprint arXiv:1711.05101}, 2017.

\bibitem{DD3D}
Dennis Park, Rares Ambrus, Vitor Guizilini, Jie Li, and Adrien Gaidon.
\newblock Is pseudo-lidar needed for monocular 3d object detection?
\newblock In {\em Proceedings of the IEEE/CVF International Conference on
  Computer Vision}, pages 3142--3152, 2021.

\bibitem{solofusion}
Jinhyung Park, Chenfeng Xu, Shijia Yang, Kurt Keutzer, Kris Kitani, Masayoshi
  Tomizuka, and Wei Zhan.
\newblock Time will tell: New outlooks and a baseline for temporal multi-view
  3d object detection.
\newblock {\em arXiv preprint arXiv:2210.02443}, 2022.

\bibitem{philion2020lift}
Jonah Philion and Sanja Fidler.
\newblock Lift, splat, shoot: Encoding images from arbitrary camera rigs by
  implicitly unprojecting to 3d.
\newblock In {\em European Conference on Computer Vision}, pages 194--210.
  Springer, 2020.

\bibitem{attention}
Ashish Vaswani, Noam Shazeer, Niki Parmar, Jakob Uszkoreit, Llion Jones,
  Aidan~N Gomez, {\L}ukasz Kaiser, and Illia Polosukhin.
\newblock Attention is all you need.
\newblock {\em Advances in neural information processing systems}, 30, 2017.

\bibitem{streampetr}
Shihao Wang, Yingfei Liu, Tiancai Wang, Ying Li, and Xiangyu Zhang.
\newblock Exploring object-centric temporal modeling for efficient multi-view
  3d object detection.
\newblock {\em arXiv preprint arXiv:2303.11926}, 2023.

\bibitem{wang2021fcos3d}
Tai Wang, Xinge Zhu, Jiangmiao Pang, and Dahua Lin.
\newblock Fcos3d: Fully convolutional one-stage monocular 3d object detection.
\newblock In {\em Proceedings of the IEEE/CVF International Conference on
  Computer Vision}, pages 913--922, 2021.

\bibitem{wang2022internimage}
Wenhai Wang, Jifeng Dai, Zhe Chen, Zhenhang Huang, Zhiqi Li, Xizhou Zhu,
  Xiaowei Hu, Tong Lu, Lewei Lu, Hongsheng Li, et~al.
\newblock Internimage: Exploring large-scale vision foundation models with
  deformable convolutions.
\newblock {\em arXiv preprint arXiv:2211.05778}, 2022.

\bibitem{wang2022detr3d}
Yue Wang, Vitor~Campagnolo Guizilini, Tianyuan Zhang, Yilun Wang, Hang Zhao,
  and Justin Solomon.
\newblock Detr3d: 3d object detection from multi-view images via 3d-to-2d
  queries.
\newblock In {\em Conference on Robot Learning}, pages 180--191. PMLR, 2022.

\bibitem{weng2019monocularpseudo}
Xinshuo Weng and Kris Kitani.
\newblock Monocular 3d object detection with pseudo-lidar point cloud.
\newblock In {\em Proceedings of the IEEE/CVF International Conference on
  Computer Vision Workshops}, pages 0--0, 2019.

\bibitem{bevformerv2}
Chenyu Yang, Yuntao Chen, Hao Tian, Chenxin Tao, Xizhou Zhu, Zhaoxiang Zhang,
  Gao Huang, Hongyang Li, Yu Qiao, Lewei Lu, et~al.
\newblock Bevformer v2: Adapting modern image backbones to bird's-eye-view
  recognition via perspective supervision.
\newblock {\em arXiv preprint arXiv:2211.10439}, 2022.

\bibitem{mvsnet}
Yao Yao, Zixin Luo, Shiwei Li, Tian Fang, and Long Quan.
\newblock Mvsnet: Depth inference for unstructured multi-view stereo.
\newblock In {\em Proceedings of the European conference on computer vision
  (ECCV)}, pages 767--783, 2018.

\bibitem{fastmvs}
Zehao Yu and Shenghua Gao.
\newblock Fast-mvsnet: Sparse-to-dense multi-view stereo with learned
  propagation and gauss-newton refinement.
\newblock In {\em Proceedings of the IEEE/CVF Conference on Computer Vision and
  Pattern Recognition}, pages 1949--1958, 2020.

\bibitem{zhu2019cbgs}
Benjin Zhu, Zhengkai Jiang, Xiangxin Zhou, Zeming Li, and Gang Yu.
\newblock Class-balanced grouping and sampling for point cloud 3d object
  detection.
\newblock {\em arXiv preprint arXiv:1908.09492}, 2019.

\bibitem{hop}
Zhuofan Zong, Dongzhi Jiang, Guanglu Song, Zeyue Xue, Jingyong Su, Hongsheng
  Li, and Yu Liu.
\newblock Temporal enhanced training of multi-view 3d object detector via
  historical object prediction.
\newblock {\em arXiv preprint arXiv:2304.00967}, 2023.

\end{thebibliography}
}


\end{document}